# COVIBOT: A Smart Chatbot for Assistance and E-Awareness during COVID-19 Pandemic

Maha Driss
*SEL Lab, CCIS, Prince Sultan University, KSA*
*RIADI Laboratory, University of Manouba, Tunisia*
mdriss@psu.edu.sa

Iman Almomani
*SEL Lab, CCIS, Prince Sultan University, KSA*
*CS Department, The University of Jordan, Amman, Jordan*
imomani@psu.edu.sa

Leen Alahmadi
*IS Department, CCSE, Taibah University, KSA*
leen.alahmadi@taibahu.edu.sa

Linah Alhajjam
*IS Department, CCSE, Taibah University, KSA*
llenafahad@taibahu.edu.sa

Raghad Alharbi
*IS Department, CCSE, Taibah University, KSA*
Raghad_khalid@taibahu.edu.sa

Shahad Alanazi
*IS Department, CCSE, Taibah University, KSA*
sha-hed@taibahu.edu.sa

*Abstract*— The coronavirus pandemic has spread over the past two years in our highly connected and information-dense society. Nonetheless, disseminating accurate and up-to-date information on the spread of this pandemic remains a challenge. In this context, opting for a solution based on conversational artificial intelligence, also known under the name of the chatbot, is proving to be an unavoidable solution, especially since it has already shown its effectiveness in fighting the coronavirus crisis in several countries. This work proposes to design and implement a smart chatbot on the theme of COVID-19, called COVIBOT, which will be useful in the context of Saudi Arabia. COVIBOT is a generative-based contextual chatbot, which is built using machine learning APIs that are offered by the cloud-based Azure Cognitive Services. Two versions of COVIBOT are offered: English and Arabic versions. Use cases of COVIBOT are tested and validated using a scenario-based approach.

## I. Introduction

COVID-19 was recognized as a pandemic in March 2020 by the World Health Organization (WHO) [1]. The spread of COVID-19 has increased rapidly around the world. Various symptoms of the virus began to appear in many people all over the world. Symptoms of Covid-19 are similar to symptoms of other illnesses and can lead to serious difficulties in people who are immunocompromised, the elderly, and those suffering from chronic illnesses [2]. As a result, people started to get furious with this virus especially as several false pieces of information about COVID-19 were shared on social media, which put pressure on hospitals and medical call centers. For example, the health center "937" of the Ministry of Health in Saudi Arabia (MOH) received more than 1.400.000 calls within one month about the emerging coronavirus during March 2020 [3]. To combat fake news and disinformation, several countries created chatbots that offer reliable information about COVID-19 such as Aapka Chikitsak, the Indian chatbot [4], and ANA, the Brazilian Portuguese chatbot [5]. Additionally, several recently published research works have opted for the use of chatbot technology and have proved its efficiency in providing telehealth services in the context of the COVID-19 pandemic [6-10] or for various purposes related to the healthcare field such as medical diagnosis, treatment of medication and pathology queries, or even for general medical awareness purposes [11-12]. To be part of these solutions, we propose to design and implement a smart chatbot, named COVIBOT that is customized for the Saudi Arabia context. The role of this chatbot is to ensure assistance and spread e-awareness. COVIBOT is built on Machine Learning (ML) algorithms [13] that are deployed to measure the similarity between user queries and the various responses pulled from trusted information sources with the aim to provide correct and accurate responses. Our proposed chatbot is a usable digital solution that can be utilized by users of different ages and intellectual backgrounds. It provides round-the-clock services and allows to save time and effort regarding COVID-19's information screening. The main functionalities offered by COVIBOT are: 1) Provide a rapid diagnosis to people with various symptoms, 2) Keep people informed of the latest news related to the COVID-19 pandemic, 3) Offer mental support to patients during and after the quarantine period, and 4) Guide people on how to change their behavior to avoid getting infected with COVID-19.

This paper is organized as follows. Section II describes the architecture, the requirement analysis, and the general and detailed design of COVIBOT. Implementation and validation details are discussed in Section III. Finally, section IV summarizes the main contributions made in this work and outlines directions for future research.

## II. COVIBOT: Architecture, Requirement Analysis, and System Design

In this paper, we propose to develop a smart chatbot, which will be useful for offering assistance and spreading COVID-19 e-awareness in Saudi Arabia. ML techniques will be used to correctly and accurately answer user questions. In the following subsections, we provide the analysis and design of the proposed chatbot, COVIBOT.

### A. Architecture

Our chatbot operates by following three successive phases, which are [14-16]: 1) the connecting phase, 2) the understanding phase, and 3) the responding phase. Fig.1. illustrates the architecture of our proposed chatbot.

In the connecting phase, the user sends a query in the form of plain text via the user interface. As a result, the user gets a response to his/her query.

The understanding phase allows processing the user's request by using: 1) the Natural Language Understanding (NLU) module, 2) the Knowledge Base (KB) module, and 3) the Dialog Manager (DM) module.

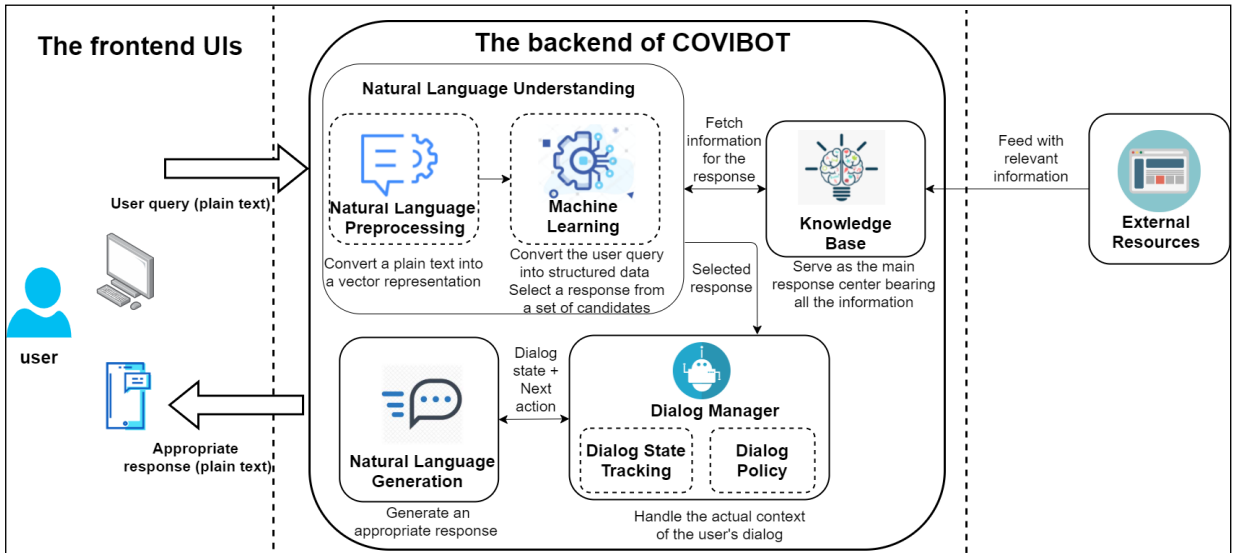

Fig. 1. COVIBOT's architecture.

NLU helps understand users' input/query, classify intents, and extract entities. It includes two sub-modules that run sequentially: Natural Language Preprocessing (NLP) and ML sub-modules. NLP submodule allows to clean the input text data and encode it into a numeric vector representation to be manipulated by ML algorithms. The role of the ML submodule is to convert the user query into structured data and select a response from a set of candidates. KB is the "brain" of the chatbot. This is where the chatbot can find the resources and information it needs to make decisions. The primary role of the KB is to save the correspondences that exist between any keywords/phrases used in the user's inputs/queries and the most adequate responses that will be provided by the chatbot. NLP helps the chatbot understand the user's queries and ML algorithms allow to find adequate and accurate responses to these queries while using the KB that has been enriched from external reputable information resources. The role of DM is to manage the state of the last context of the user's dialog. For instance, let's assume that the user enters the following query: "What are the preventative measures for COVID-19?", so the chatbot will consider this query and start processing it. After a while, the user changes his mind and re-enters a new query, which is the following one: "What are the symptoms of COVID-19?". In this case, the user refers in his last query to the symptoms of COVID-19 and not to the preventive measures. Thus, the chatbot must correctly understand the last context of the dialog and make the necessary changes to properly handle the user's query. This is ensured by DM. DM has two submodules: the Dialogue State Tracking (DST) and the Dialogue Policy (DP). DST keeps the last state of the user's dialog and uses the corresponding input intent. It defines whether the new received entity values should modify the existing entity values. DP allows training the chatbot to take the next smart action. DM interacts with DP to decide the next best action to take. DP can be used to add a related follow-up question or provide suggestions to enrich the dialogue or let the user confirm a chosen answer.

The responding phase includes the Natural language Generation (NLG) module, where the chatbot responses are constructed. It does the reverse of NLU and converts structured data to text. Additionally, NLG encloses a set of pre-defined template messages that represent action names. Thus, based on the action specified by DM, the corresponding template message is invoked. If the template in question requires certain values to be fulfilled, these values are transmitted by the DM to the NLG. In the end, the most suitable text message will be displayed to the user.

*B. Requirement Analysis*

To specify users' requirements that are fulfilled by our proposed chatbot, we use the use case diagram, which is an effective modeling and specification technique offered by Unified Modeling Language (UML) [17]. Fig. 2. illustrates the use case diagram that shows the interaction between the actors and different functionalities offered by our chatbot. The principal actor is the user. Two secondary actors feed the chatbot's KB with reliable information, namely: the web applications of the Saudi Ministry of Health and the WHO. The main functions provided by our chatbot are: 1) "Provide diagnosis check", 2) "Get guidance", 3) "View FAQ", 4) "Provide mental support for contaminated people", 5) "Deliver information", and 6) "View vaccines". These use cases are linked with other sub-use cases by different relationships, which are: "include", "extend", and generalization. For example, "Provide diagnosis check" includes "Ask for user symptoms confirmation", which is extended by "Suggest COVID-19 test" in case the user shows severe symptoms of COVID-19.

In Table I., we provide the detailed textual description of the "Provide diagnosis check" use case. The textual descriptions are elaborated for all use cases to provide additional information for the requirement analysis phase.

As non-functional requirements, COVIBOT should fulfill the following quality properties [17]:

- **High responsiveness**: COVIBOT should be able to render responses quickly to provide a fluid conversation with the user.
- **High usability**: COVIBOT should provide simple and intuitive user interfaces to allow the user to

- invoke the offered functionalities with effectiveness, efficiency, and satisfaction.
- **High reliability**: COVIBOT's KB should be fed by accurate information extracted from reputable sources.
- **High availability**: COVIBOT should offer round-the-clock services.

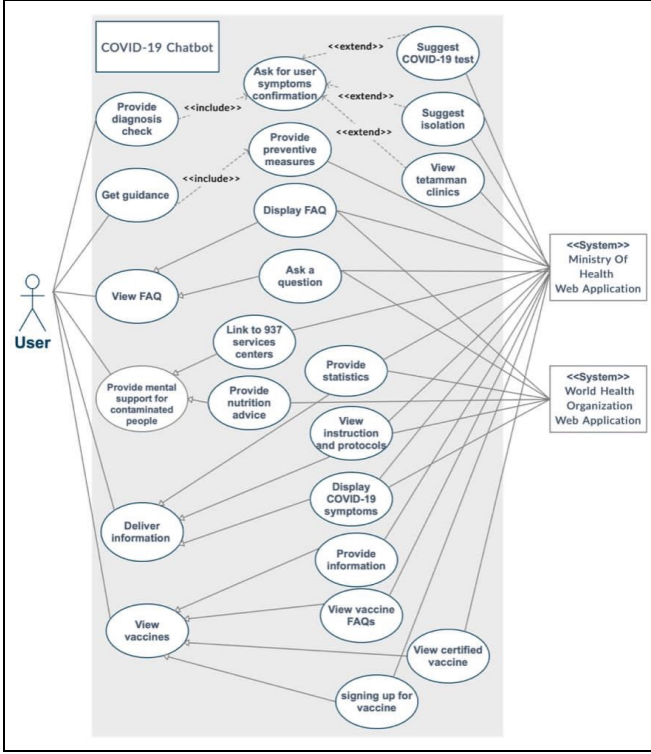

Fig. 2. Use case diagram of COVIBOT.

TABLE I. TEXTUAL DESCRIPTION OF THE "PROVIDE DIAGNOSIS CHECK" USE CASE

| Use Case Name | Provide diagnosis check. |
|---|---|
| Actor(s) | The "User" |
| Description | The user is diagnosed via chatbot. |
| Pre-condition(s) | The user sends a query to get diagnosed. |
| Post-condition(s) | After receiving a request and specifying the symptoms, the chatbot assesses the severity of the user's symptoms. |
| Flow of Events | 1. The user visits the chatbot main page;<br>2. The user sends a query by selecting the "Provide diagnosis check" function;<br>3. The chatbot asks questions and gives options to select for the precise specification of the symptoms;<br>4. The chatbot evaluates the severity of the user's symptoms. |
| Alternative Flows | None. |

### C. System Design

As a general design for our proposed chatbot, we used a component diagram, which is a UML diagram that aims to illustrate the structural relationships between the logical system components. Fig. 3. provides the COVIBOT's component diagram. It presents the interactions that exist between the User Interfaces (UIs) and the different modules involved in the understanding and responding phases of the chatbot's operation.

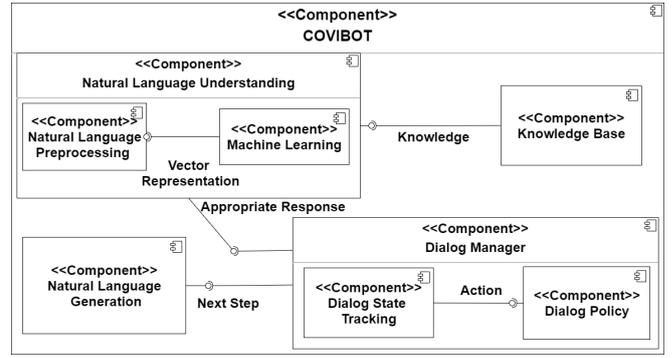

Fig. 3. Component diagram of COVIBOT.

To provide a detailed design specification, we illustrate in Fig. 4 the flowchart of the main component of COVIBOT, which is NLU [4]. The NLU flowchart starts with a user input message that goes into two processes: Entity Recognition (ER) and Intent Classification (IC). EC allows the analysis of the text to identify the topic the user asked about. It recognizes the entities by looking for similar categories of words, required information, or any other data entered by the user. IC permits to classify the user intents. The outputs of ER and IC are entities and intents, respectively. The entity represents the core idea of the user's request and the intent is the action to be performed by the chatbot in response to the user's request. DM and KB are involved in the process of candidate response generation. The candidate response generator performs domain-related semantic calculations to correctly process the user's query. A list of candidate responses is provided as a result. The response selector uses the dialog's context as well as the intent of the entities extracted from the last user's query to select the most adequate response.

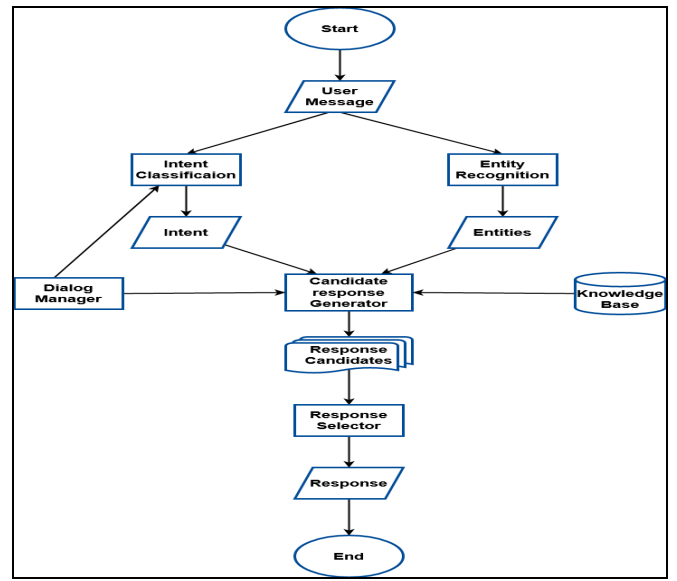

Fig. 4. Flowchart of NLU module.

### III. IMPLEMENTATION AND VALIDATION

COVIBOT was implemented using a set of cognitive services provided by the cloud platform Azure [18-19]. These services bring artificial intelligence capabilities within the reach of developers and without the need for ML expertise. You can simply call an API to integrate the ability to find, understand, and accelerate decision-making in

software applications. Among these services, we used for the implementation of COVIBOT:

- **Language Understanding (LUIS)** [20]: is a conversational AI cloud service that draws on the intelligence provided by ML algorithms and natural language processing techniques to understand the meaning, as well as to extract relevant and detailed related information from users queries. In our work, we used LUIS to implement the NLU and KB modules. A diverse set of algorithms are offered by the cloud platform Azure, including multiclass decision forest, recommendation systems, neural network regression, multiclass neural network, and K-means clustering. In our work, two ML algorithms are used, which are: decision forest regression and K-means clustering. The high performance provided by these algorithms in NL processing applications justifies our choice [21];
- **QnA Maker** [22]: is a cloud service that uses natural language processing techniques to generate a natural conversational layer from the data that are exchanged. It allows finding the most suitable answers for the entries of the KB. In our work, we used QnA Maker to implement the NLU and KB modules;
- **Bot Framework Composer** [23]: is an open-source visual authoring canvas used to implement bots. It offers several language understanding services that offer advanced composition of bot replies using advanced language generation techniques. This framework encloses various value-added features that can be used to build an advanced conversational experience such as a visual editing canvas for dialog flow, a powerful language generation and templating system, and a tool to create and manage language understanding. In our work, we used the Bot Framework Composer for constructing the dialogue flow.

To feed up the KB module, accurate information about the COVID-19 pandemic in Saudi Arabia was extracted from two reliable sources, which are the web applications of the Saudi Ministry of Health and the WHO as indicated previously in the use case diagram of COVIBOT. First, once signed up in the QnAMaker, the URLs of our trusted sources were added. Then, all the information collected was saved, trained, and tested to ensure that it has been added successfully. Finally, KB was published to connect it with Bot Framework Composer. The KB information is updated on a regular basis, usually every 1-2 weeks.

To implement the NLU module, LUIS and Bot Framework Composer were used. When the user asks a question, LUIS helps the bot to understand the user's intent and reply with the best answer (i.e., entity) that has the highest similarity score among candidate responses. To generate intents and entities, we used either the LUIS portal or the Bot Framework Composer by assigning user utterances to the intents. As a result, a set of intents and entities is built, trained, and published to be used by NLU. The resulting set of intents is added to the BOT Framework Composer as triggers, which are composed of phrases having similar meanings with the associated intents. Fig. 6. shows the phrases triggering the "Guidance" intent. These phrases have a similarity score of 0.7 with the associated intent.

After collecting intents and entities, dialogues are built and deployed using the BoT Framework Composer. Fig. 6. illustrates the construction of the "Diagnosis Check" dialog. The resulting dialogues are saved in JSON format. They are tested and validated using emulations that are ensured by the local bot runtime manager of the Bot Framework Composer.

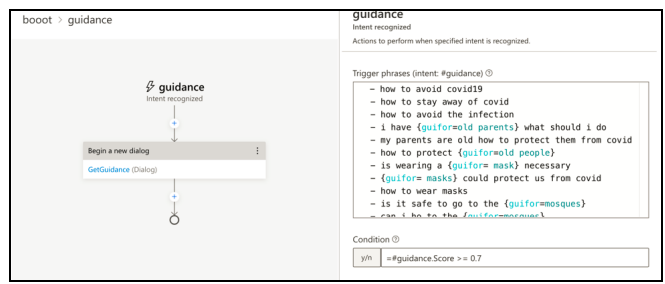

Fig. 5. Trigger phrases of the "Guidance" intent.

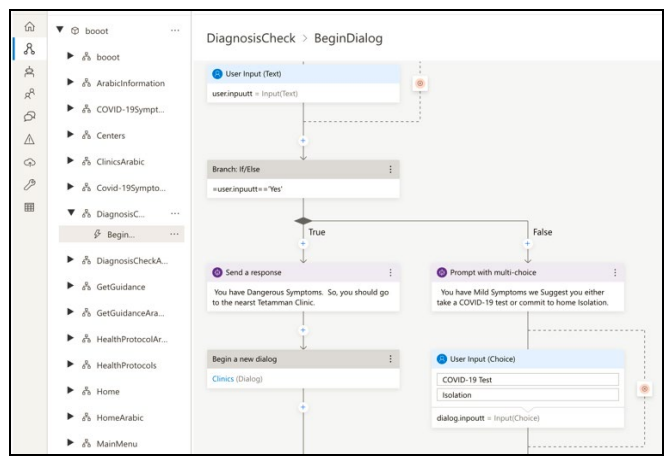

Fig. 6. "Diagnosis Check" dialog constructed using the Bot Framework Composer.

To test and validate the functional and non-functional requirements related to the implementation of COVIBOT, we opt for a scenario-based approach [24]. For the sake of concision and clarity, two scenarios are considered in this paper: one using the English version and the other one using the Arabic version.

In the first scenario, *Noura* wants to protect herself and her family from COVID-19. She used COVIBOT to learn about COVID-19 vaccines. Once she started the conversation with COVIBOT, a welcoming and introduction card with the language option is displayed as illustrated in Fig. 7. *Noura* selected the English language. A list of options she can choose from appeared. But this time, *Noura* typed in the conversation "Is there a vaccine for covid19?" as shown by Fig. 8. A card responding to her query appeared along with a list of options about COVID-19 vaccines. So, she selected "Certified Vaccine". In response, a card with a "Learn More" button that allows opening a pdf file containing details about certified vaccines in Saudi Arabia was displayed. She clicked "Learn More" and the pdf file opened in her browser as illustrated by Fig. 9. After consulting the provided document, *Noura* wanted to register to get vaccines' doses and COVIBOT suggested to her to register in the "Sehhaty" application as shown by Fig. 10.

In the second scenario, *Fawaz* chose to use the Arabic version. He used the Saudi dialect to query COVIBOT, which in turn responded using the same dialect. For the first time, *Fawaz* asked about a diagnosis check as illustrated by Fig. 11. A list of symptoms is then displayed by COVIBOT. *Fawaz* was confirmed to have severe symptoms as shown by Fig. 12. So, COVIBOT suggested visiting urgently the nearest "Tatamman" clinic. *Fawaz* chose Assir's health care center and in response COVIBOT displayed the location of this center in Google Maps as illustrated by Fig. 13.

COVIBOT has shown correct and precise operation through the invocation of the different functionalities that have been suggested in the two scenarios. Extensive testing experiments using a scenario-based approach have also been conducted. The testing experiments have proved the usability and the effectiveness of the overall functionalities provided by COVIBOT.

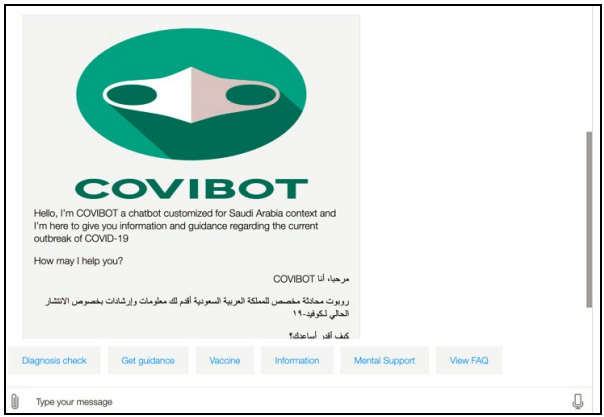

Fig. 7.  English version: Welcoming and introduction page of COVIBOT.

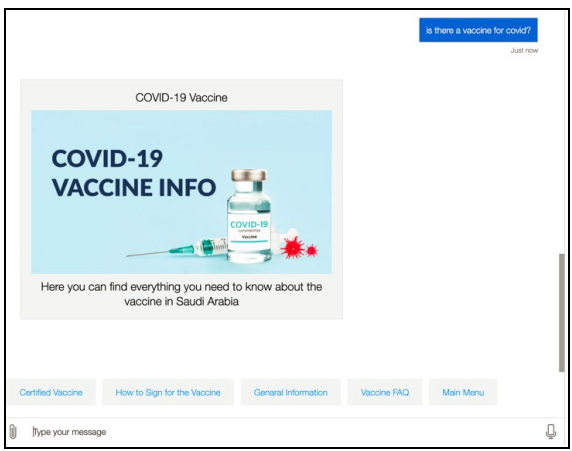

Fig. 8.  English version: Vaccine card.

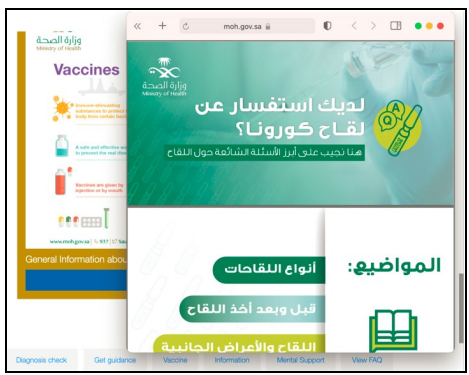

Fig. 9.  PDF file containing information about certified vaccines in Saudi Arabia provided by the Ministry of Health.

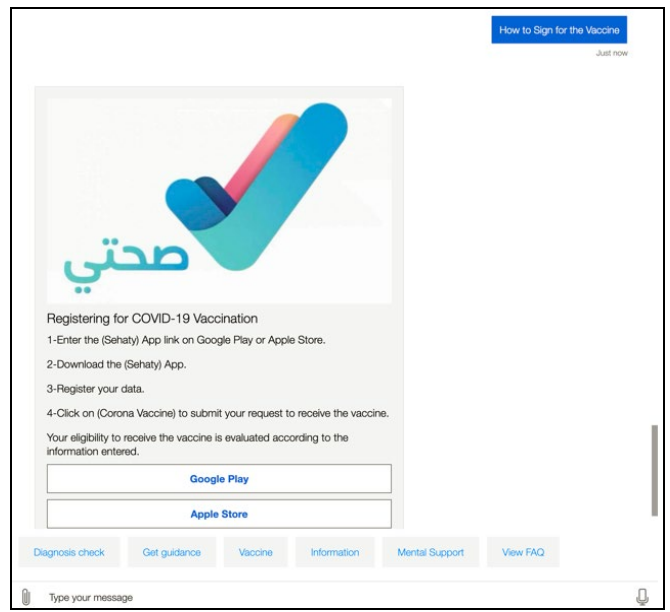

Fig. 10. English version: Suggestion to use "Sehhaty" application to receive vaccines' doses.

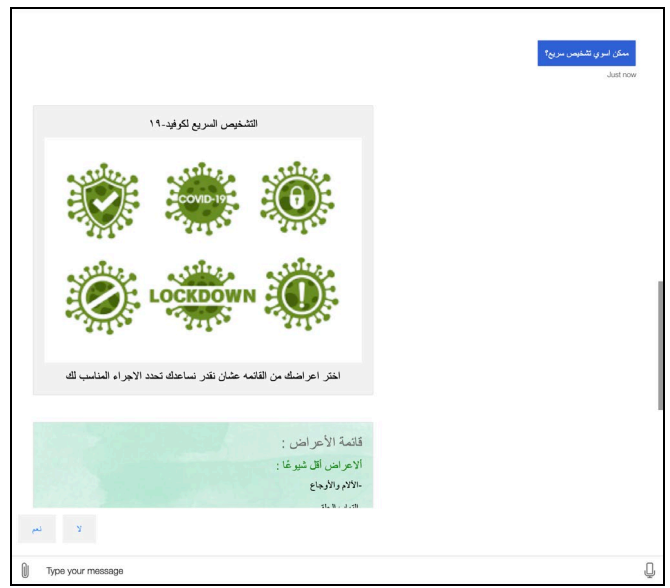

Fig. 11. Arabic version: Diagnosis check.

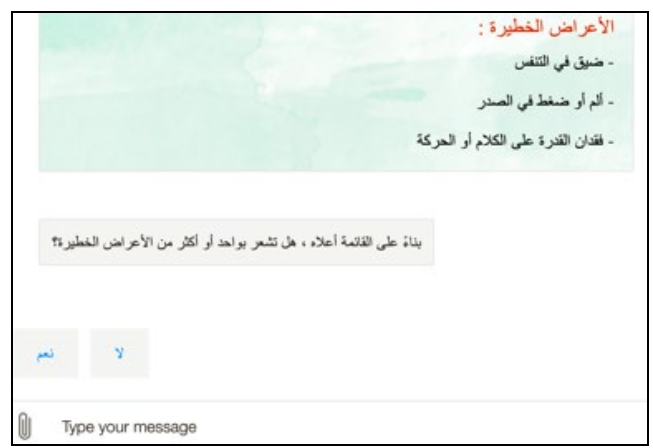

Fig. 12. Arabic version: COVID-19 symptoms' list.

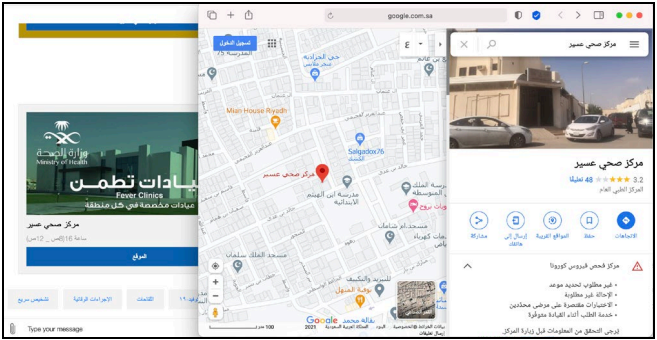

Fig. 13. Arabic version: Tatamman clinics' localization.

## IV. Conclusion and Future Work

This paper details the design and implementation of a smart chatbot application used to ensure assistance and e-awareness during the COVID-19 pandemic in Saudi Arabia. In our work, we propose to use ML techniques offered by the cognitive services of the cloud platform Azure to respond correctly to the user's queries. Our chatbot, called COVIBOT, operates by following three successive phases, which are: 1) the connecting phase, 2) the understanding phase, and 3) the responding phase. In the connecting phase, the user sends a query in the form of plain text via the user interface. The understanding phase allows processing the user query. The responding phase is responsible for building the chatbot responses. Two versions of COVIBOT are implemented and tested, an English version and an Arabic version. As future work, we intend to conduct an analytical performance comparison between our application and existing COVID-19 applications and chatbots. We are also planning to enhance our chatbot by considering the following extensions: 1) add a speech recognition module, 2) provide online doctors' consultations, and 3) make COVIBOT a generic software solution capable to provide assistance and e-awareness also for the most common diseases in Saudi Arabia.